\documentclass[runningheads]{llncs}

\usepackage[T1]{fontenc}
\usepackage{graphicx}
\usepackage{xcolor}
\usepackage{subcaption}

\newcommand{\am}{\textit{ASSISTments}}
\newcommand{\arc}{\textit{ARC}}
\newcommand{\race}{\textit{RACE++}}
\newcommand{\racefourk}{\textit{R.4k}}
\newcommand{\raceeightk}{\textit{R.8k}}
\newcommand{\racetwelvek}{\textit{R.12k}}

\newcommand{\ie}{i.e., }
\newcommand{\qdet}{QDET}
\newcommand{\qo}{Q\textsubscript{O}} 
\newcommand{\qc}{Q\textsubscript{C}} 
\newcommand{\qa}{Q\textsubscript{A}} 
\newcommand{\wv}{word2vec}
\newcommand{\cg}{\textcolor{lightgray}}

\begin{document}

\title{A quantitative study of NLP approaches to question difficulty estimation\thanks{Submitted to AIED 2023 (24th International Conference on Artificial Intelligence in Education). This paper reports on research supported by Cambridge University Press \& Assessment. We thank Dr Andrew Caines for the feedback on the manuscript.}}
\titlerunning{A quantitative study of NLP approaches to question difficulty estimation}

\author{
Luca Benedetto\inst{1}\orcidID{0000-0002-5113-4696} 
}

\authorrunning{L. Benedetto}

\institute{
Dept. of Computer Science \& Technology, University of Cambridge, United Kingdom\\
\email{luca.benedetto@cl.cam.ac.uk}\\
}

\maketitle

\begin{abstract}
Recent years witnessed an increase in the amount of research on the task of Question Difficulty Estimation from Text (\qdet{}) with Natural Language Processing (NLP) techniques, with the goal of targeting the limitations of traditional approaches to question calibration.
However, almost the entirety of previous research focused on single \textit{silos}, without performing quantitative comparisons between different models or across datasets from different educational domains.
In this work, we aim at filling this gap, by quantitatively analyzing several approaches proposed in previous research, and comparing their performance on three publicly available real world datasets containing questions of different types from different educational domains.
Specifically, we consider reading comprehension Multiple Choice Questions (MCQs), science MCQs, and math questions.
We find that Transformer based models are the best performing across different educational domains, with DistilBERT performing almost as well as BERT, and that they outperform other approaches even on smaller datasets.
As for the other models, the hybrid ones often outperform the ones based on a single type of features, the ones based on linguistic features perform well on reading comprehension questions, while frequency based features (TF-IDF) and word embeddings (word2vec) perform better in domain knowledge assessment.

\end{abstract}

\section{Introduction}\label{sec:intro}
Estimating the difficulty of exam questions is an important part of any educational setup.
Indeed, it enables to identify questions of low quality or that are not suitable for a certain group of learners, as well as to perform custom recommendations. 
Pretesting and manual calibration are the traditional approaches to difficulty estimation and have been used for decades, but they are time-consuming and expensive. 
In recent years, Question Difficulty Estimation from Text (\qdet{}) by means of NLP techniques has become a fairly popular research topic, as a way to target the limitations of traditional approaches to question calibration.
Indeed, text is always available at the time of question creation and, if we were able to estimate question difficulty from it, we could overcome (or at least reduce) the need for pretesting and manual calibration.
Most of previous works focus on specific \textit{silos}, without performing comparisons between different approaches, or across datasets from different educational domains, or with questions of different type.
In this work, we perform a quantitative evaluation of approaches proposed in previous research for \qdet{}, and compare their performance on three publicly available real world datasets from different educational domains, with the goal of understanding if and how the accuracy of different models varies across different types of questions.
Specifically, we experiment on i) reading comprehension MCQs (\race{}), ii) science MCQs (\arc{}), and iii) math questions (\am{}); all the questions are in English.
We find that Transformer based models are the best performing across different educational domains (DistilBERT performing almost as well as BERT), even on smaller datasets.
As for the other models, the hybrid ones often outperform the ones based on a single type of features, the ones based on linguistic features perform well on reading comprehension questions, while frequency based features (TF-IDF) and word embeddings (word2vec) perform better in domain knowledge assessment.
Code and supplementary material is available at \url{https://github.com/lucabenedetto/qdet-comparison}.

\section{Related Works}
Whilst research on question difficulty estimation has a fairly long history, it was only in recent years that NLP approaches gained popularity in this task.
A recent survey \cite{alkhuzaey2021systematic} analyzed some high level trends of the domain, and another \cite{benedetto2022survey} described previous approaches and proposed a taxonomy to categorize them.
However, neither performed a quantitative experimental evaluation.

Previous research focused on both the language assessment domain - which aims at evaluating the language proficiency of learners - and the content knowledge assessment domain - which assesses the knowledge about a specific topic.
Four papers focused on reading \textit{comprehension} MCQs, which are specific to language assessment and are made of three components: a reading passage (or context), a stem, and a certain number of answer options, one of them being correct. 
The passage heavily affects question difficulty, and indeed previous \qdet{} models leveraged it.
In \cite{huang2018development}, the authors directly use reading difficulty as a proxy of question difficulty.
In \cite{bi2021simple}, the authors compute five linguistic features from the text of the question and the text of the passage, and compare their value with a threshold to obtain one of two difficulty levels.
In \cite{huang2017question} the authors estimate the difficulty with \wv{} embeddings \cite{mikolov2013distributed} and a fully-connected neural network.
Lastly, in \cite{lin2019automated} the authors explicitly take into consideration the relation between the reading passage and the question, by using an attention mechanism \cite{vaswani2017attention} to model the relevance of each sentence in the reading passage.

The rest of the previous literature focused on \textit{knowledge} questions, which do not have an accompanying passage and appear both in language assessment and domain knowledge assessment. 
Most of the proposed approaches have a clear distinction between a feature engineering phase and a regression phase, and only a minority uses end-to-end-neural networks.
Considering the first group, the most commonly used machine learning regression algorithms are Random Forests \cite{benedetto2020r2de,benedetto2020introducing,yaneva2019predicting,yaneva2020predicting},
SVM \cite{ehara2018building,yang2018feature,lee2019manipulating,hsu2018automated}, and 
linear regression \cite{el2017predicting,hou2019modeling,trace2017determining}, with some works also experimenting with weighted softmax \cite{settles2020machine} and ridge regression \cite{pandarova2019predicting}.
As for the features, a large variety of approaches have been proposed: linguistic features \cite{hou2019modeling,trace2017determining}, \wv{} embeddings \cite{ehara2018building,hsu2018automated}, frequency-based features \cite{benedetto2020r2de}, and readability indexes.
Also, multiple papers experimented with hybrid approaches created by combining some of the aforementioned features: 
linguistic features and \wv{} embeddings \cite{yang2018feature},
linguistic and frequency-based features \cite{settles2020machine},
frequency-based and linguistic features and readability indexes \cite{benedetto2020introducing}, 
word embeddings and linguistic features and frequency-based features \cite{yaneva2019predicting,yaneva2020predicting}.

Previous approaches that use end-to-end neural networks mostly use Transformers and the attention mechanism. 
Specifically, the attention mechanism is used in \cite{cheng2019dirt,qiu2019question}, BERT \cite{devlin2019bert} is used in \cite{zhou2020multi,benedetto2021application,tong2020exercise}, 
and DistilBERT \cite{sanh2019distilbert} in \cite{benedetto2021application}.

\section{Evaluated Models}
We experiment with i) linguistic features, ii) readability indexes, iii) TF-IDF (Term Frequency - Inverse Document Frequency), iv) \wv{} embeddings, v) several hybrid approaches, and vi) Transformers.
%

\textbf{Linguistic features} have been used in multiple forms in previous research \cite{culligan2015comparison,hou2019modeling,trace2017determining}.
They are measures related to the number and length of words and sentences in the question, the answer choices (for MCQs) and the context (for reading comprehension questions).
We use seventeen linguistic features, taking them from previous research, and use them as input to a Random Forest regression model.

\textbf{Readability indexes} are measures designed to evaluate how easy a reading passage is to understand, and have been used for \qdet{} in \cite{huang2018development}.
Following the examples of previous research, we experiment with: 
Flesch Reading Ease \cite{flesch1948new}, 
Flesch-Kincaid Grade Level \cite{kincaid1975derivation}, 
ARI \cite{senter1967automated}, 
Gunning FOG Index \cite{gunning1952technique}, 
Coleman-Liau Index \cite{coleman1965understanding},
Linsear Write Formula \cite{klare1974assessing},
and Dale-Chall Readability Score \cite{dale1948formula}.
%
We use them as input features to a Random Forest regression model.

Frequency-based features have been used in \cite{settles2020machine,benedetto2020r2de}, and we use \textbf{TF-IDF} \cite{manning2008introduction}.
The TF-IDF weights represent how important is a word (or a set of words) to a document in a corpus: the importance grows with the number of occurrences of the word in the document but it is limited by its frequency in the whole corpus; intuitively, words that are very frequent in all the documents of the corpus are not important to any of them.
Following the example of previous research, we consider three approaches to encode the questions:
i) \qo{} considers only the question,
ii) \qc{} appends the text of the correct option to the question,
iii) \qa{} concatenates all the options (both correct and wrong) to the question; \qc{} and \qa{} can be used only on MCQs.
The TF-IDF features are then used as input to a Random Forest regression model.

\textbf{Word2vec} \cite{mikolov2013distributed} has been the most common technique for building word embeddings in previous research \cite{ehara2018building}, therefore this is the non-contextualized word embedding technique we evaluate.
We experiment with the same three approaches to create the embeddings as with TF-IDF (\qo{}, \qc{}, and \qa{}), and use the \wv{} features as input to a Random Forest regression model. 

\textbf{Hybrid Approaches} were also used in previous research, and they are all obtained by concatenating features from two (or more) of the approaches presented above, and using them as input to a single Random Forest regression model.
Specifically, we evaluate
i) linguistic and readability features \cite{beinborn2015candidate,benedetto2020introducing,lee2019manipulating},
ii) linguistic, readability, and TF-IDF \cite{benedetto2020introducing},
iii) linguistic features and word embeddings \cite{yang2018feature},
iv) linguistic features, TF-IDF, and word embeddings \cite{yaneva2019predicting,yaneva2020predicting}.

\textbf{Transformers} are attention-based pre-trained language models that can be fine-tuned to target various downstream tasks.
This generally leads to better performance with shorter training times with respect to training the neural model from scratch, because it leverages the pre-existing knowledge of the pre-trained model.
Attention-based models gained huge popularity in recent years and \qdet{} was no exception \cite{benedetto2021application,zhou2020multi,fang2019exercise,huang2017question,tong2020exercise}.
Following the examples of \cite{benedetto2021application,zhou2020multi}, we experiment with \textbf{BERT} \cite{devlin2019bert} and \textbf{DistilBERT} \cite{sanh2019distilbert}, fine-tuning the publicly available pre-trained models on the task of \qdet{}.
Again, we evaluate the same three approaches for encoding: \qo{}, \qc{}, and \qa{}.

\section{Experimental Datasets}
We briefly introduce here the datasets that are used in this study.

\textbf{\race{}} is a dataset of reading comprehension MCQs, built by merging the original \textit{RACE} dataset \cite{lai2017race}, with the \textit{RACE-c} dataset \cite{liang2019new}.
Each question is made of a reading passage, a stem, and four possible answer options, one of them being correct.
Each question is manually assigned one out of three difficulty levels (0, 1, 2), which we consider as gold standard for \qdet{}.
The dataset is not perfectly balanced, with 1 being the most frequent label.
The \textit{train} split contains 100,568 questions from 27,572 reading passages,
and \textit{test} contains 5,642 questions from 1,542 passages; there are no reading passages shared across splits.
Since \race{} is a very large dataset for \qdet{}, we also build three smaller versions of its \textit{train} split (\racefourk{}, \raceeightk{}, and \racetwelvek{}), which are obtained by randomly sampling 4000, 8000, and 12000 questions respectively; \textit{test} remains the same as in the original dataset.
We use them to study how the models performance varies depending on the size of the training dataset.

\textbf{\arc{}} \cite{clark2018think} is a dataset of science MCQs, each being assigned a level between 3 and 9, which we use as gold standard for \qdet{}.
The original dataset contains questions with a varying number of answer choices but, for consistency with \race{}, we only keep items with three distractors and one correct option; the resulting \textit{train}, and \textit{test} splits contain 3,358 and 3,530 questions respectively.
\arc{} is very unbalanced: the two most common labels (8 and 5) appear about 1400 and 700 times respectively, the least common (6) around 100.
Thus, we partially balance its \textit{train} split, by randomly subsampling the two most common labels to keep only 500 questions for each of them; all the details and results shown here are for the balanced dataset.

\textbf{\am{}} \cite{feng2009addressing} is a dataset of questions mostly about math; differently from the other datasets, no answer options are available.
Questions are not manually calibrated but instead pretested, and the difficulty is continuously distributed in the range $[-5; 5]$ in a Gaussian-like shape, with $\mu=-0.32$ and $\sigma=1.45$.
The \textit{train} split contains 6,736 questions, and \textit{test} 2,245.

\section{Results}\label{sec:results}

\subsection{Comparison with ground truth difficulty}
\qdet{} is a sentence regression task, commonly evaluated by comparing the estimated values with the gold standard references.
Here, we use the metrics that are most common in the literature\footnote{
The F1 score is also common in previous research; we do not use it because it does not take into consideration the order of difficulty levels, thus it is not ideal for \qdet{}.
}: Root Mean Squared Error (RMSE), R2 score, and Spearman's $\rho$.
To study how stable each model is, we perform five independent training runs and show the mean and standard deviation of the metrics on the \textit{test} set.
It is important to remark here that although the difficulty in \race{} and \arc{} are discrete levels, all the \qdet{} models are trained as regression models and output a continuous difficulty, which we convert to one of the discrete values with simple thresholds (\ie{} mapping to the closest label).

\begin{table}
\centering
\begin{tabular}{l | c c c } 
Model & RMSE & R2 & Spearman's $\rho{}$ \\ 
\hline
Random                			& 1.026 \cg{$\pm$0.004} & -1.899 \cg{$\pm$0.022} & -0.005 \cg{$\pm$0.011} \\ 
Majority              			& 0.616 \cg{$\pm$0.000} & -0.046 \cg{$\pm$0.000} & - \\ 
\hline
Ling.           	  			& 0.471 \cg{$\pm$0.004} & 0.388 \cg{$\pm$0.011} & 0.653 \cg{$\pm$0.005} \\ 
Read.       			    	& 0.552 \cg{$\pm$0.003} & 0.160 \cg{$\pm$0.008} & 0.507 \cg{$\pm$0.005} \\ 
W2V Q\textsubscript{Only} (\qo{})   			    	& 0.513 \cg{$\pm$0.002} & 0.276 \cg{$\pm$0.005} & 0.570 \cg{$\pm$0.003} \\ 
W2V Q\textsubscript{Correct} (\qc{})   			    	& 0.518 \cg{$\pm$0.002} & 0.263 \cg{$\pm$0.006} & 0.559 \cg{$\pm$0.003} \\ 
W2V Q\textsubscript{All} (\qa{})  				    	& 0.507 \cg{$\pm$0.005} & 0.291 \cg{$\pm$0.013} & 0.580 \cg{$\pm$0.008} \\ 
TF-IDF \qo{}            			& 0.516 \cg{$\pm$0.013} & 0.265 \cg{$\pm$0.038} & 0.568 \cg{$\pm$0.025} \\ 
TF-IDF \qc{}            			& 0.508 \cg{$\pm$0.005} & 0.290 \cg{$\pm$0.015} & 0.585 \cg{$\pm$0.010} \\ 
TF-IDF \qa{}            			& 0.511 \cg{$\pm$0.011} & 0.280 \cg{$\pm$0.031} & 0.577 \cg{$\pm$0.022} \\ 
\hline
Ling., Read.      				& 0.478 \cg{$\pm$0.002} & 0.371 \cg{$\pm$0.006} & 0.644 \cg{$\pm$0.003} \\ 
W2V \qa{}, Ling. 				& 0.467 \cg{$\pm$0.003} & 0.399 \cg{$\pm$0.008} & 0.658 \cg{$\pm$0.005} \\ 
Ling., Read., TF-IDF \qc{} 		& 0.463 \cg{$\pm$0.002} & 0.409 \cg{$\pm$0.005} & 0.668 \cg{$\pm$0.004} \\ 
W2V \qa{}, Ling., TF-IDF \qc{}	& 0.464 \cg{$\pm$0.003} & 0.408 \cg{$\pm$0.004} & 0.666 \cg{$\pm$0.005} \\ 
\hline
DistilBERT \qo{}      			& 0.391 \cg{$\pm$0.009} & 0.578 \cg{$\pm$0.019} & 0.778 \cg{$\pm$0.009} \\ 
DistilBERT \qc{}      			& 0.391 \cg{$\pm$0.007} & 0.579 \cg{$\pm$0.015} & 0.777 \cg{$\pm$0.007} \\ 
DistilBERT \qa{}      			& 0.381 \cg{$\pm$0.009} & 0.600 \cg{$\pm$0.019} & \textbf{0.790} \cg{$\pm$0.007} \\ 
BERT \qo{}            			& 0.383 \cg{$\pm$0.008} & 0.597 \cg{$\pm$0.021} & 0.778 \cg{$\pm$0.011} \\ 
BERT \qc{}            			& 0.410 \cg{$\pm$0.007} & 0.537 \cg{$\pm$0.018} & 0.752 \cg{$\pm$0.008} \\ 
BERT \qa{}            			& \textbf{0.372} \cg{$\pm$0.012} & \textbf{0.619} \cg{$\pm$0.028} & 0.789 \cg{$\pm$0.014} \\ 
\end{tabular}
\caption{Evaluation of the models on \race{}.}
\label{tab:eval_racepp}
\end{table}

Table \ref{tab:eval_racepp} presents the results for \race{}.
The table is divided into four parts: i) random and majority baselines, ii) linguistic, readability, \wv{} and TF-IDF features, iii) hybridy approaches, and iv) Transformers.
All the models outperform the two baselines and Transformers are significantly better than the others according to all metrics; most likely, the attention can capture the relations between the passage and the question.
However, there is not a clear advantage of BERT over DistilBERT.
For Transformers, using the text of the answer choices (\qa{}) seems to improve the accuracy of the estimation, but the difference with \qo{} and \qc{} is often minor.
As for the other features, the Linguistic perform better than Readability, \wv{}, and TF-IDF.
Whilst it makes sense that \wv{} and TF-IDF are not very effective here, as \race{} is a language assessment dataset, the fact that the Readability features perform poorly partially comes as a surprise.
Considering \wv{} and TF-IDF, in most cases we can see no significant difference between the results obtained with the three different encodings (\qo{}, \qc{}, and \qa{}).
Hybrid models seem to bring some advantages: most of the combinations outperform the single features, and using a larger number of different features leads to greater advantages.

\begin{table}[ht!]
\centering
\begin{tabular}{l | c c c } 
Model & RMSE & R2 & Spearman's $\rho{}$ \\ 
\hline
Random                			& 2.726 \cg{$\pm$0.025} & -1.473 \cg{$\pm$0.046} & 0.002 \cg{$\pm$0.017} \\ 
Majority              			& 2.195 \cg{$\pm$0.000} & -0.603 \cg{$\pm$0.000} & - \\ 
\hline
Ling.	             			& 1.633 \cg{$\pm$0.007} & 0.113 \cg{$\pm$0.007} & 0.333 \cg{$\pm$0.012} \\ 
Read.	             			& 1.612 \cg{$\pm$0.002} & 0.135 \cg{$\pm$0.002} & 0.363 \cg{$\pm$0.003} \\ 
W2V \qo{}	        			& 1.695 \cg{$\pm$0.005} & 0.044 \cg{$\pm$0.006} & 0.223 \cg{$\pm$0.011} \\ 
W2V \qc{}	        			& 1.686 \cg{$\pm$0.006} & 0.054 \cg{$\pm$0.007} & 0.240 \cg{$\pm$0.012} \\ 
W2V \qa{}	        			& 1.707 \cg{$\pm$0.002} & 0.030 \cg{$\pm$0.002} & 0.203 \cg{$\pm$0.004} \\ 
TF-IDF \qo{}            			& 1.644 \cg{$\pm$0.009} & 0.100 \cg{$\pm$0.010} & 0.322 \cg{$\pm$0.010} \\ 
TF-IDF \qc{}            			& 1.642 \cg{$\pm$0.006} & 0.102 \cg{$\pm$0.006} & 0.320 \cg{$\pm$0.011} \\ 
TF-IDF \qa{}            			& 1.643 \cg{$\pm$0.010} & 0.101 \cg{$\pm$0.011} & 0.322 \cg{$\pm$0.014} \\ 
\hline
Ling., Read.	      			& 1.592 \cg{$\pm$0.004} & 0.157 \cg{$\pm$0.004} & 0.392 \cg{$\pm$0.003} \\ 
W2V \qc{}, Ling.	 			& 1.619 \cg{$\pm$0.004} & 0.128 \cg{$\pm$0.004} & 0.354 \cg{$\pm$0.003} \\ 
Ling., Read., TF-IDF \qa{}		& 1.602 \cg{$\pm$0.003} & 0.146 \cg{$\pm$0.003} & 0.377 \cg{$\pm$0.003} \\ 
W2V \qc{}, Ling., TF-IDF \qa{}	& 1.583 \cg{$\pm$0.005} & 0.166 \cg{$\pm$0.006} & 0.400 \cg{$\pm$0.006} \\ 
\hline
DistilBERT \qo{}      			& 1.755 \cg{$\pm$0.007} & -0.025 \cg{$\pm$0.008} & 0.108 \cg{$\pm$0.125} \\ 
DistilBERT \qc{}      			& 1.620 \cg{$\pm$0.115} & 0.122 \cg{$\pm$0.125} & 0.312 \cg{$\pm$0.193} \\ 
DistilBERT \qa{}      			& 1.537 \cg{$\pm$0.112} & 0.210 \cg{$\pm$0.120} & 0.430 \cg{$\pm$0.181} \\ 
BERT \qo{}            			& 1.753 \cg{$\pm$0.021} & -0.023 \cg{$\pm$0.025} & 0.107 \cg{$\pm$0.084} \\ 
BERT \qc{}            			& 1.742 \cg{$\pm$0.041} & -0.011 \cg{$\pm$0.047} & 0.132 \cg{$\pm$0.117} \\ 
BERT \qa{}            			& \textbf{1.535} \cg{$\pm$0.148} & \textbf{0.208} \cg{$\pm$0.154} & \textbf{0.502} \cg{$\pm$0.139} \\ 
\end{tabular}
\caption{Evaluation of the models on \arc{}.}
\label{tab:eval_arc}
\end{table}

The results for \arc{} are shown in Table \ref{tab:eval_arc}.
Again, all the models outperform the simple baselines, and the best performing is BERT \qa{}, followed by DistilBERT \qa{}.
A crucial difference with \race{} is that the \qa{} Transformers are the only ones that perform well, while \qc{} and \qo{} perform on part with (if not worse than) most of the other models.
Also, the Transformers are not very stable - most likely due to the limited and unbalanced number of questions per difficulty level - as can be seen from the standard deviation of the evaluation metrics, suggesting that a lot of care should be put on hyper-parameters at training time.
Considering the other models, there is not one type of feature which is clearly better than the others, but again hybrid models bring some advantages (\wv{} \qc{} + Ling. + TF-IDF \qa{} is the third best approach).

The results for \am{} are shown in Table \ref{tab:eval_am}.
Since the difficulty is a continuous value in the range $[-5; +5]$, we use as baseline the average difficulty of \textit{train} instead of the majority label.
\begin{table}[ht!]
\centering
\begin{tabular}{l | c c c } 
Model                 		& RMSE            & R2               & Spearman's $\rho{}$   \\ 
\hline
Random                			& 3.222 \cg{$\pm$ 0.037} & -4.248 \cg{$\pm$ 0.119} & 0.010 \cg{$\pm$ 0.020} \\ 
Average                			& 1.408 \cg{$\pm$0.000} & -0.002 \cg{$\pm$0.000} & - \\ 
\hline
Ling.	             			& 1.380 \cg{$\pm$ 0.002} & 0.038 \cg{$\pm$ 0.002} & 0.209 \cg{$\pm$ 0.006} \\ 
Read.	             			& 1.372 \cg{$\pm$ 0.002} & 0.049 \cg{$\pm$ 0.002} & 0.228 \cg{$\pm$ 0.013} \\ 
W2V \qo{}	        			& 1.278 \cg{$\pm$ 0.002} & 0.175 \cg{$\pm$ 0.002} & 0.375 \cg{$\pm$ 0.009} \\ 
TF-IDF \qo{} 			  			& 1.274 \cg{$\pm$ 0.004} & 0.179 \cg{$\pm$ 0.005} & 0.346 \cg{$\pm$ 0.004} \\ 
\hline
Ling., Read.	      			& 1.363 \cg{$\pm$ 0.004} & 0.061 \cg{$\pm$ 0.005} & 0.259 \cg{$\pm$ 0.010} \\ 
W2V \qo{}, Ling.	 			& 1.296 \cg{$\pm$ 0.006} & 0.151 \cg{$\pm$ 0.008} & 0.343 \cg{$\pm$ 0.013} \\ 
W2V \qo{}, Ling., TF-IDF \qo{}	& 1.289 \cg{$\pm$ 0.005} & 0.160 \cg{$\pm$ 0.006} & 0.359 \cg{$\pm$ 0.010} \\ 
\hline
DistilBERT \qo{}      			& \textbf{1.267} \cg{$\pm$ 0.009} & \textbf{0.189} \cg{$\pm$ 0.012} & 0.402 \cg{$\pm$ 0.012} \\ 
BERT \qo{}            			& 1.272 \cg{$\pm$ 0.034} & 0.182 \cg{$\pm$ 0.045} & \textbf{0.441} \cg{$\pm$ 0.013} \\ 
\end{tabular}
\caption{Evaluation of the models on \am{}.}
\label{tab:eval_am}
\end{table}

For this dataset only the text of the questions is available, thus \qo{} is the only encoding that can be used.
The Transformers are again the better performing models, neither being clearly better, and they are followed closely by TF-IDF and \wv{}.
Being this a domain knowledge assessment dataset, question difficulty mostly depends on terms and topics, thus Linguistic and Readability features do not perform well.
Also, since the hybrid approaches that we experiment with are composed of one between Ling. and Read., they are outperformed by TF-IDF and \wv{}. 

\subsection{Variation in performance depending on training dataset size}
\begin{figure}
     \centering
     \begin{subfigure}[b]{0.49\textwidth}
         \centering
    \includegraphics[width=\textwidth]{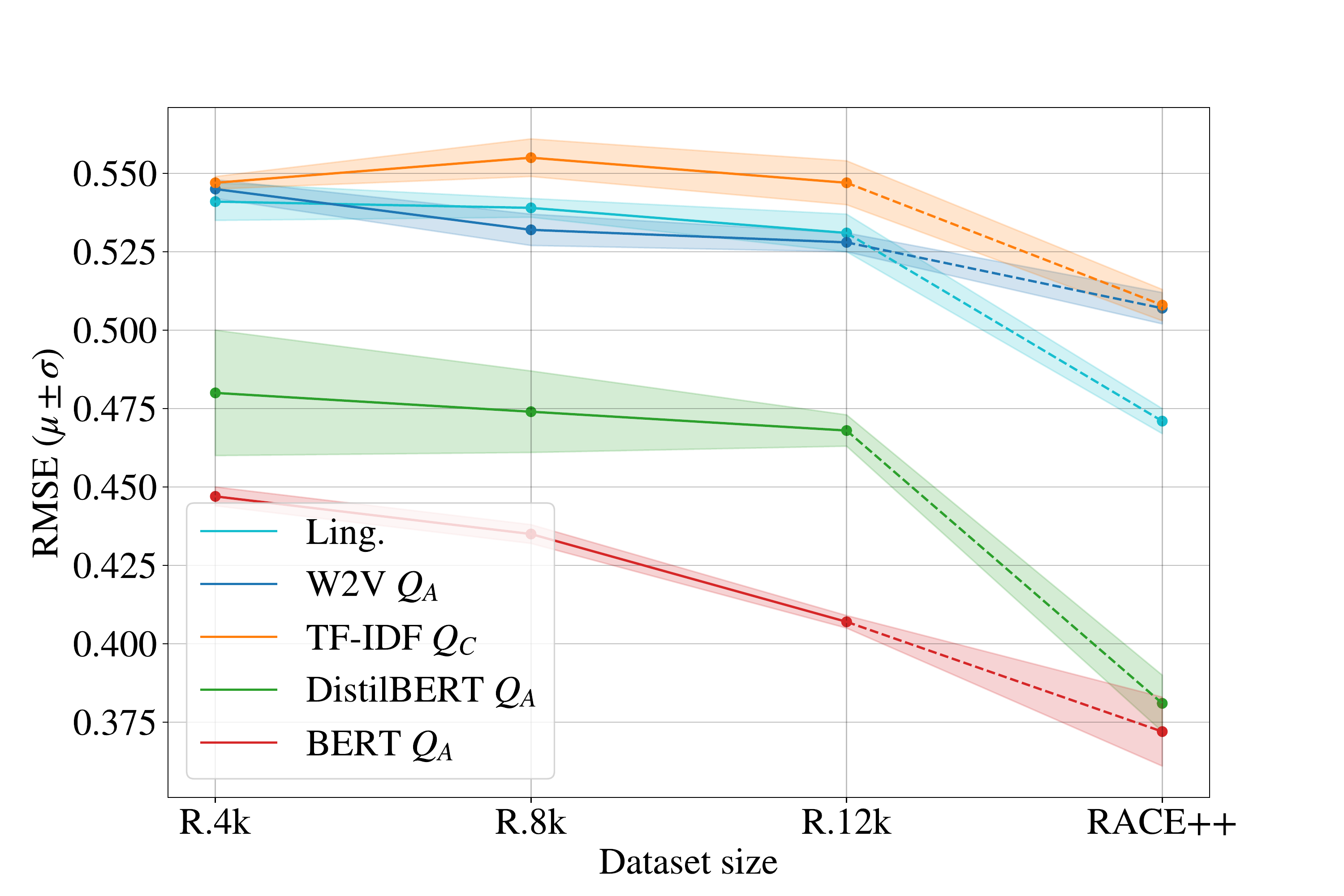}
     \end{subfigure}
     \hfill
     \begin{subfigure}[b]{0.49\textwidth}
         \centering
    \includegraphics[width=\textwidth]{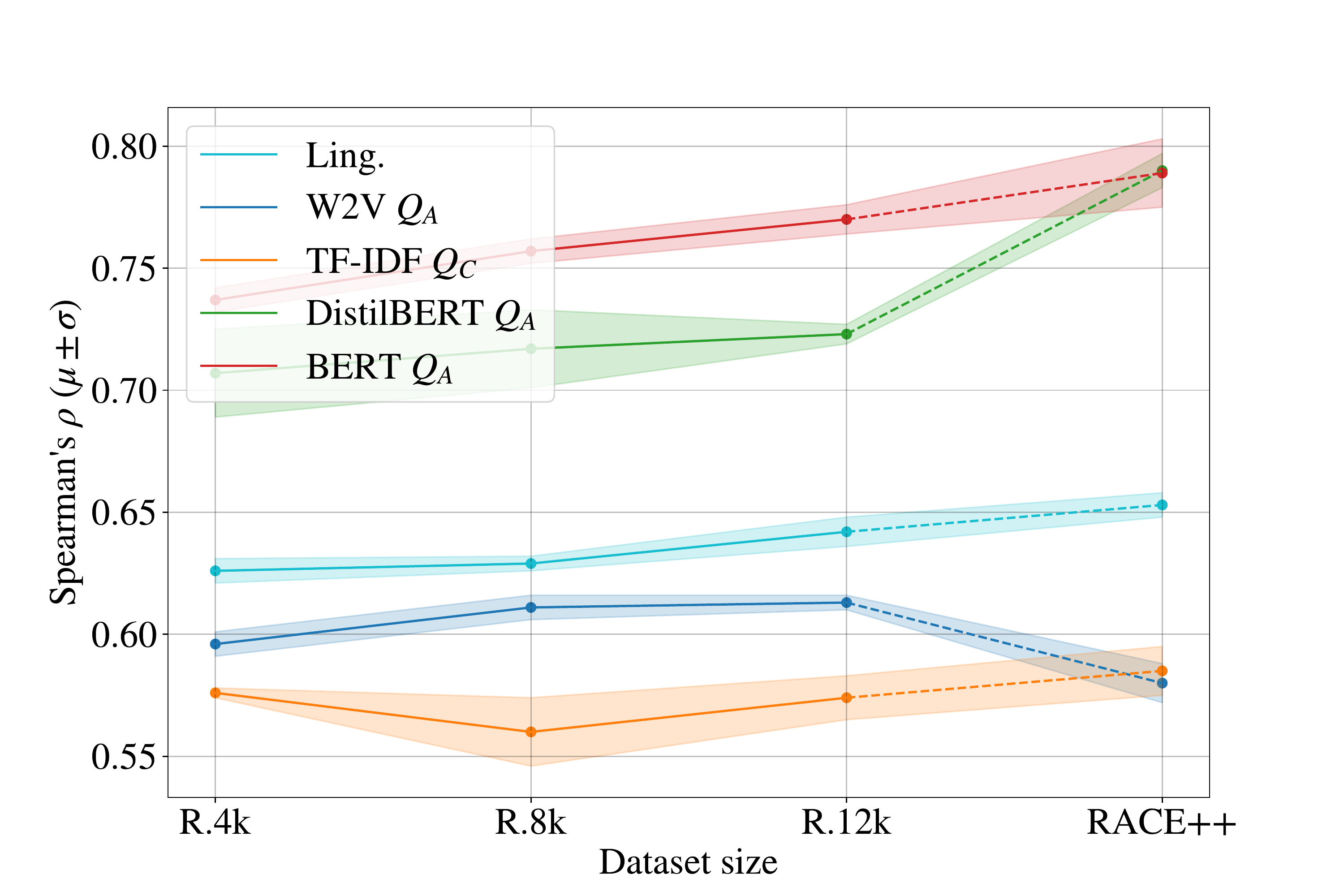}
     \end{subfigure}
        \caption{RMSE and Spearman's $\rho$ for different sizes of the training set.}
        \label{fig:metrics_per_size}
\end{figure}

To study how the model accuracy varies depending on the training set size, we built three reduced versions of the \textit{train} set of \race{}: \racefourk{}, \raceeightk{}, and \racetwelvek{}.
Figure \ref{fig:metrics_per_size} plots, for different models, how the evaluation metrics on the test set vary depending on the training size. 
We show here RMSE and Spearman's $rho$ (the mean $\pm$ the standard deviation), but the findings are similar for R2.

As expected, models tend to get better for larger training sizes, and in particular there is a significant step between \racetwelvek{} (12,000 questions) and the whole \race{} (100,000).
However, even when trained on the smallest \racefourk{}, Transformers are better than the other models trained on the whole dataset, suggesting that it might be interesting to consider even smaller training sizes.

\subsection{Distribution of estimated difficulties}\label{subsec:distr_est_diff}
\begin{figure}
     \centering
     \begin{subfigure}[b]{0.32\textwidth}
         \centering
    \includegraphics[width=\textwidth]{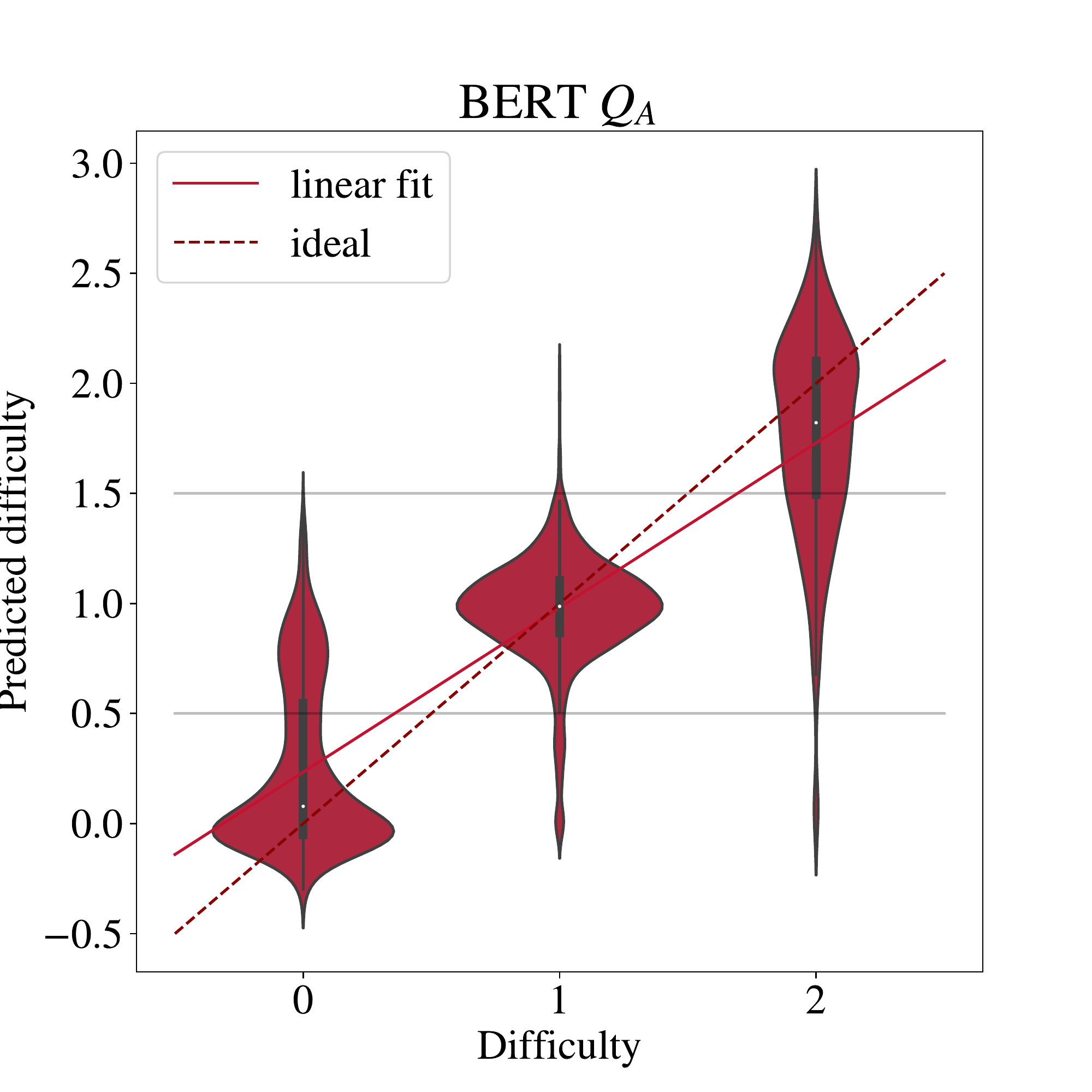}
     \end{subfigure}
     \hfill
     \begin{subfigure}[b]{0.32\textwidth}
         \centering
    \includegraphics[width=\textwidth]{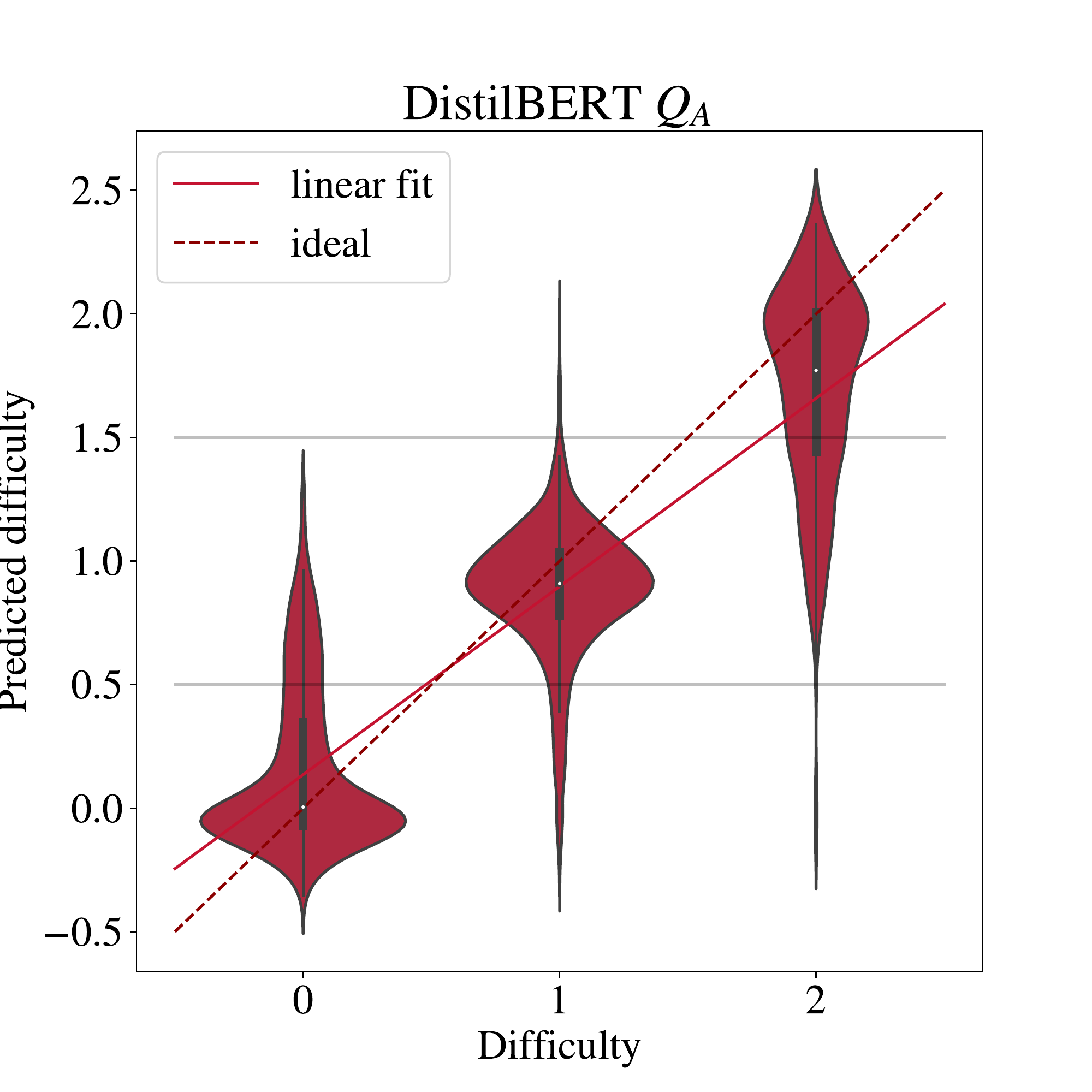}
     \end{subfigure}
     \hfill
     \begin{subfigure}[b]{0.32\textwidth}
         \centering
    \includegraphics[width=\textwidth]{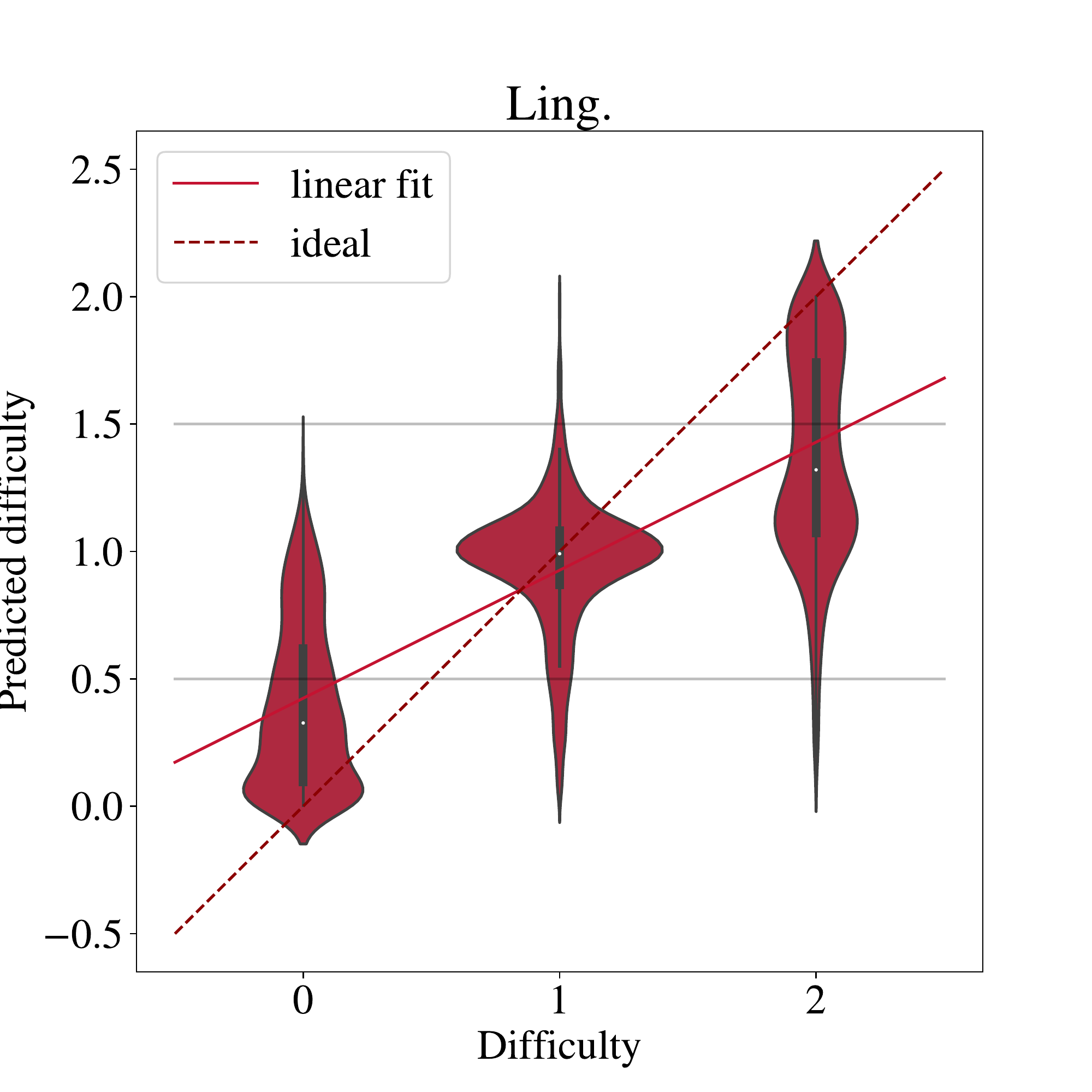}
     \end{subfigure}
       \caption{Distribution of estimated difficulties by true difficulty level, \race{}.}
       \label{fig:race_pp_distr_est_diff}
\end{figure}

To better understand how different models perform, we also study the distribution of the estimated difficulties for different difficulty levels.
Differently from the previous sections, the distribution shown here is the one before the conversion into discrete labels (for \race{} and \arc{}), and we show the thresholds as horizontal lines.
Figure \ref{fig:race_pp_distr_est_diff} plots the distribution of the difficulties estimated by the three best performing models, separately for each true difficulty level, for \race{}.
Additionally, we plot the line that linearly fits the estimated difficulties and (dashed) the ideal linear fit.

\begin{figure}[t]
     \centering
     \begin{subfigure}[b]{0.32\textwidth}
         \centering
    \includegraphics[width=\textwidth]{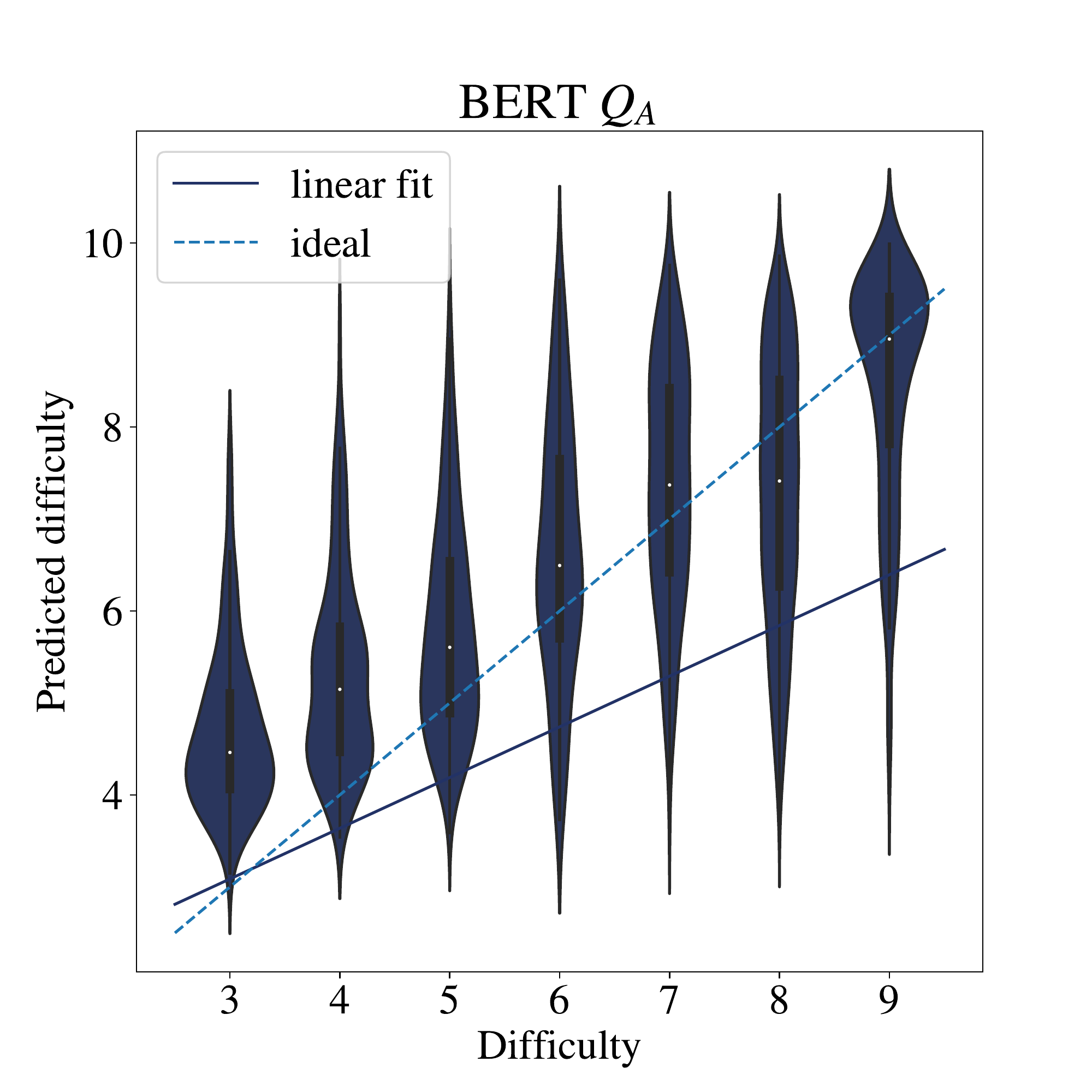}
     \end{subfigure}
     \hfill
     \begin{subfigure}[b]{0.32\textwidth}
         \centering
    \includegraphics[width=\textwidth]{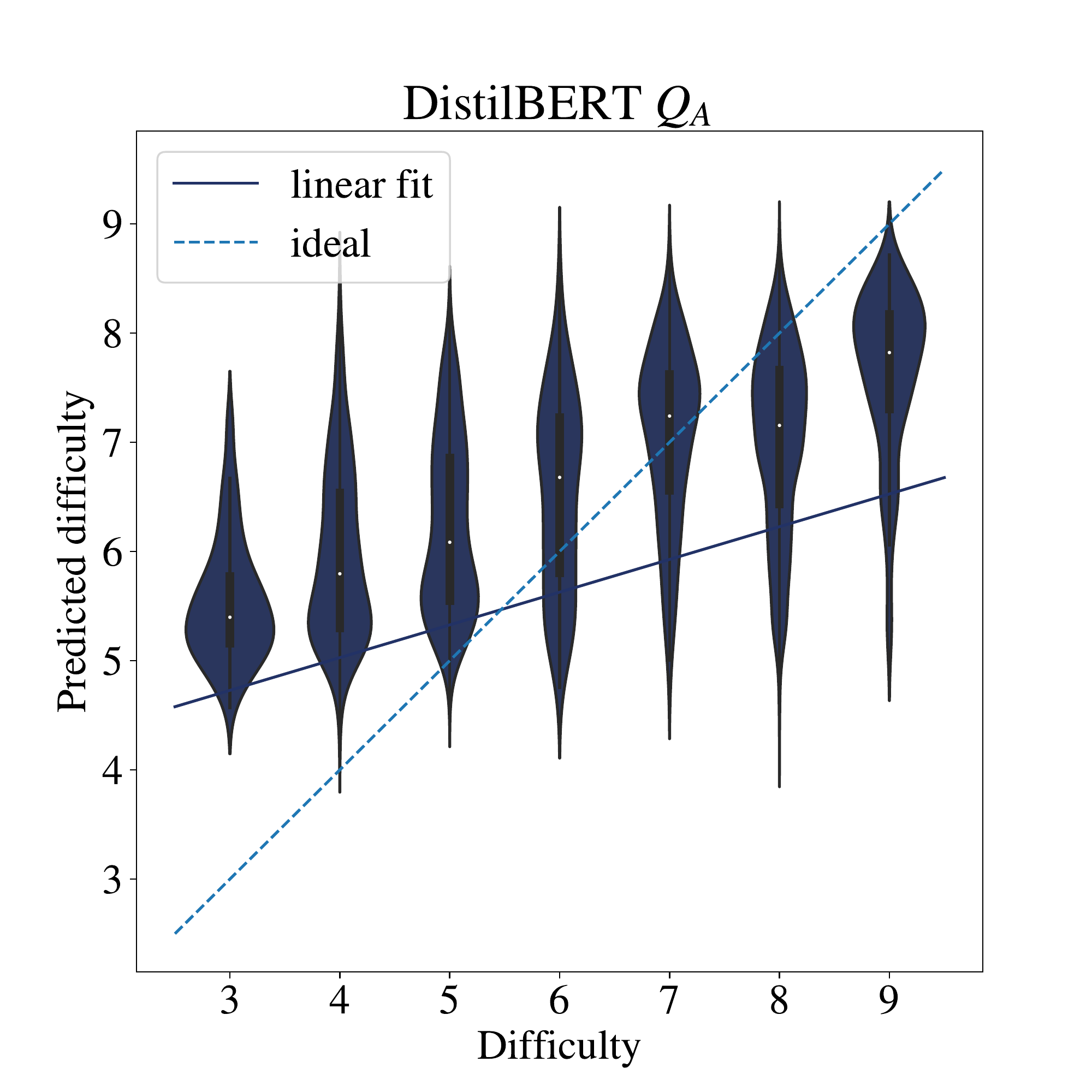}
     \end{subfigure}
     \hfill
     \begin{subfigure}[b]{0.32\textwidth}
         \centering
    \includegraphics[width=\textwidth]{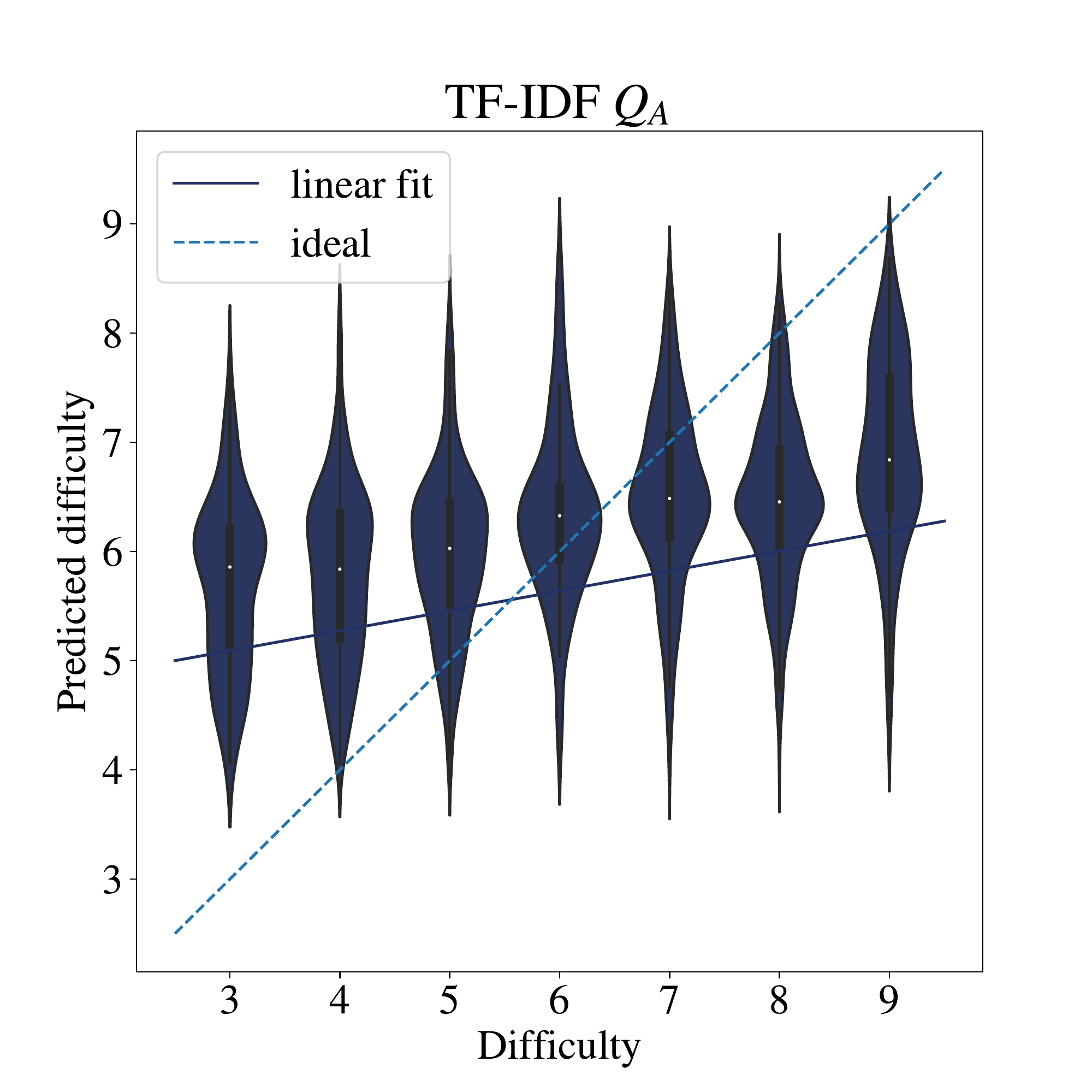}
     \end{subfigure}

       \caption{Distribution of estimated difficulties by true difficulty level, \arc{}.}
        \label{fig:arc_balanced_distr_est_diff}
\end{figure}

The figures show that the performance is almost perfect for the questions with difficulty $1$, and the majority of the difference in the accuracy is due to the other levels.
This is probably caused by the fact that \race{} is unbalanced and $1$ is the most common label, suggesting that for large datasets it might be worth performing some undersampling to reduce the imbalance in the dataset at training time.
%
Figure \ref{fig:arc_balanced_distr_est_diff} shows the same analysis for \arc{}.
It is immediately visible the difference with \race{}, and the fact that it is a more difficult dataset overall: indeed, there are more difficulty levels for the models to learn, they are unbalanced, and the number of questions is much smaller.
Still, we can see that the performance of BERT is better than the other models, especially for the high difficulty and low difficulty items, while the difficulties between 6 and 8 (included) seem to be the most challenging.

%
\begin{figure}[t]
     \centering
     \begin{subfigure}[b]{0.32\textwidth}
         \centering
    \includegraphics[width=\textwidth]{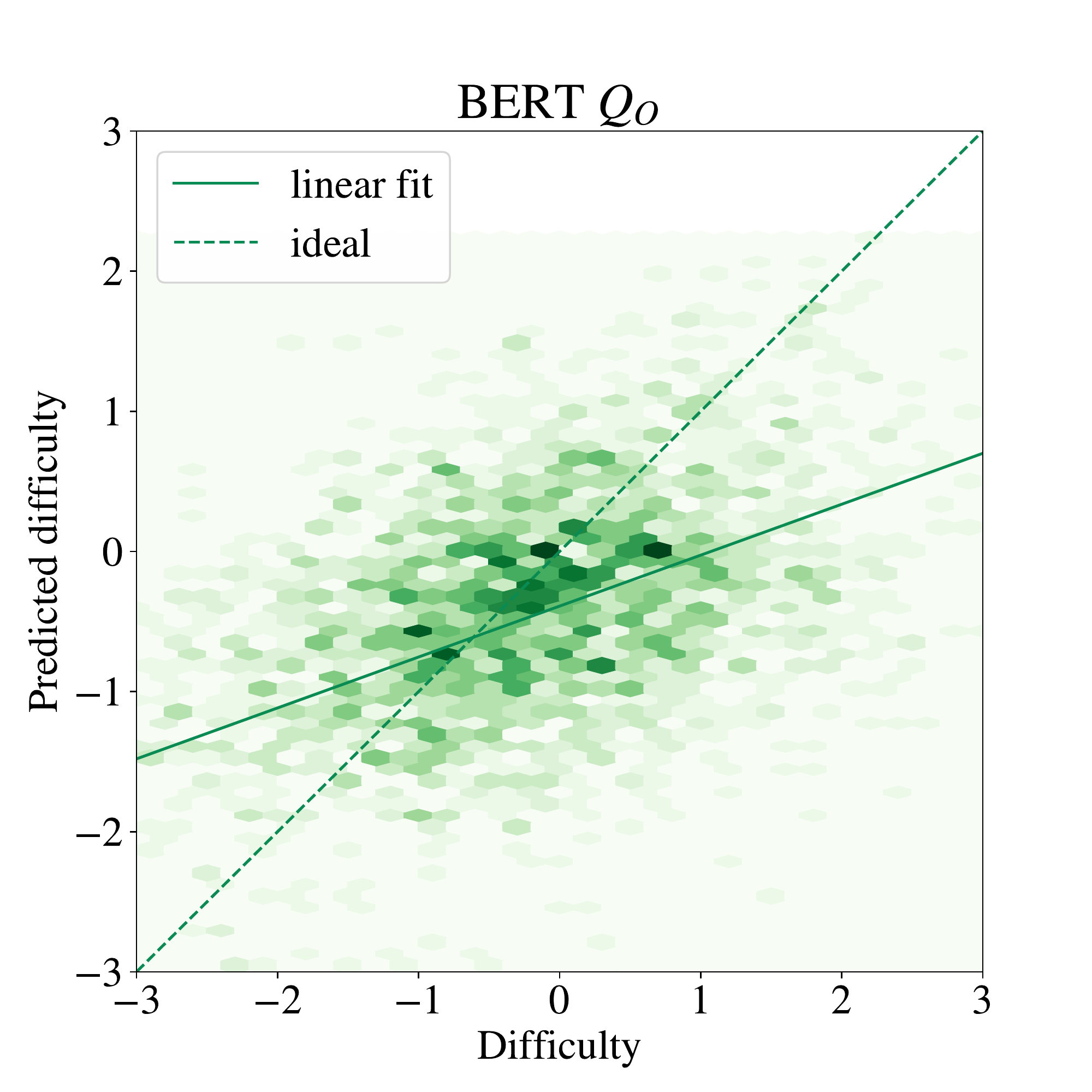}
     \end{subfigure}
     \hfill
     \begin{subfigure}[b]{0.32\textwidth}
         \centering
    \includegraphics[width=\textwidth]{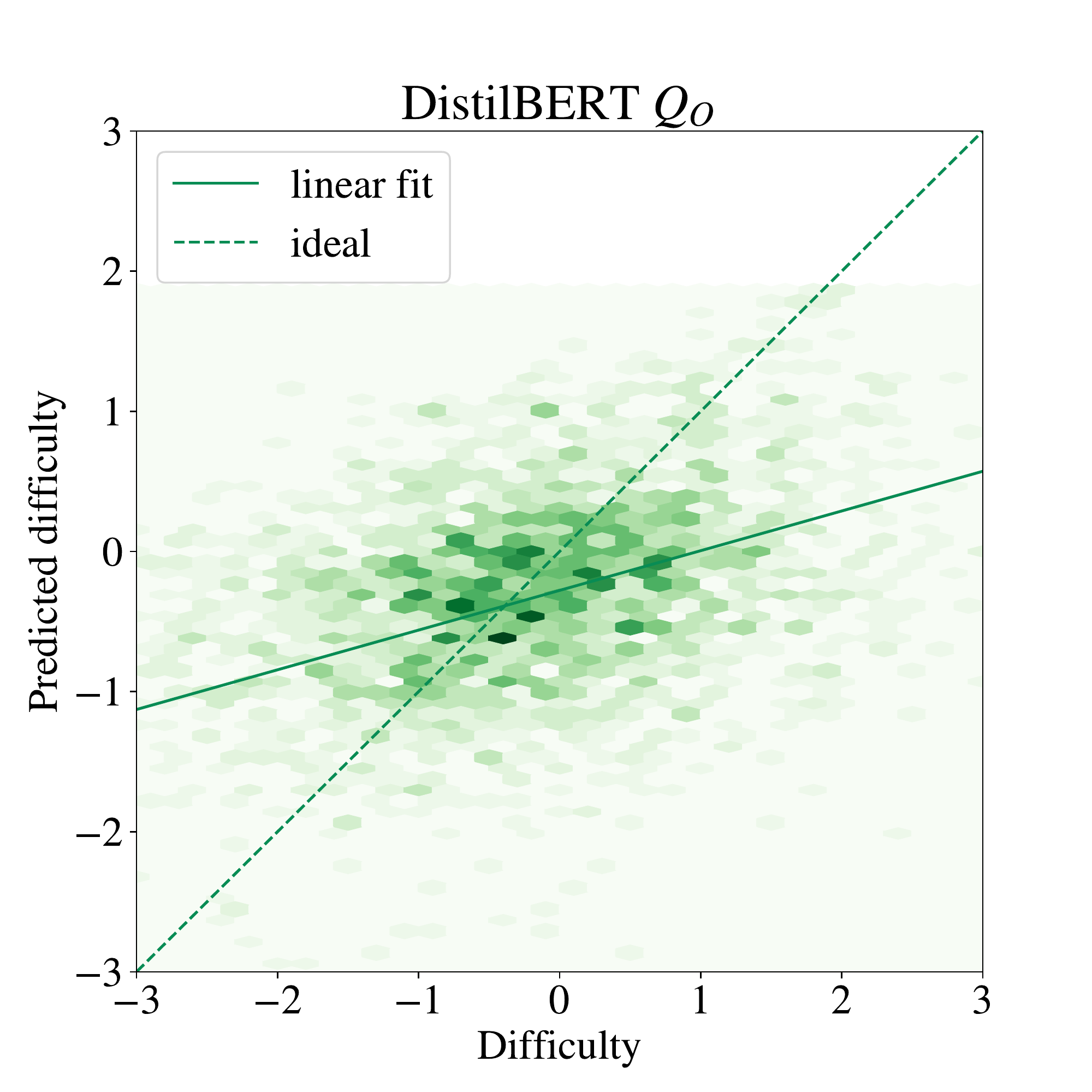}
     \end{subfigure}
          \hfill
     \begin{subfigure}[b]{0.32\textwidth}
         \centering
    \includegraphics[width=\textwidth]{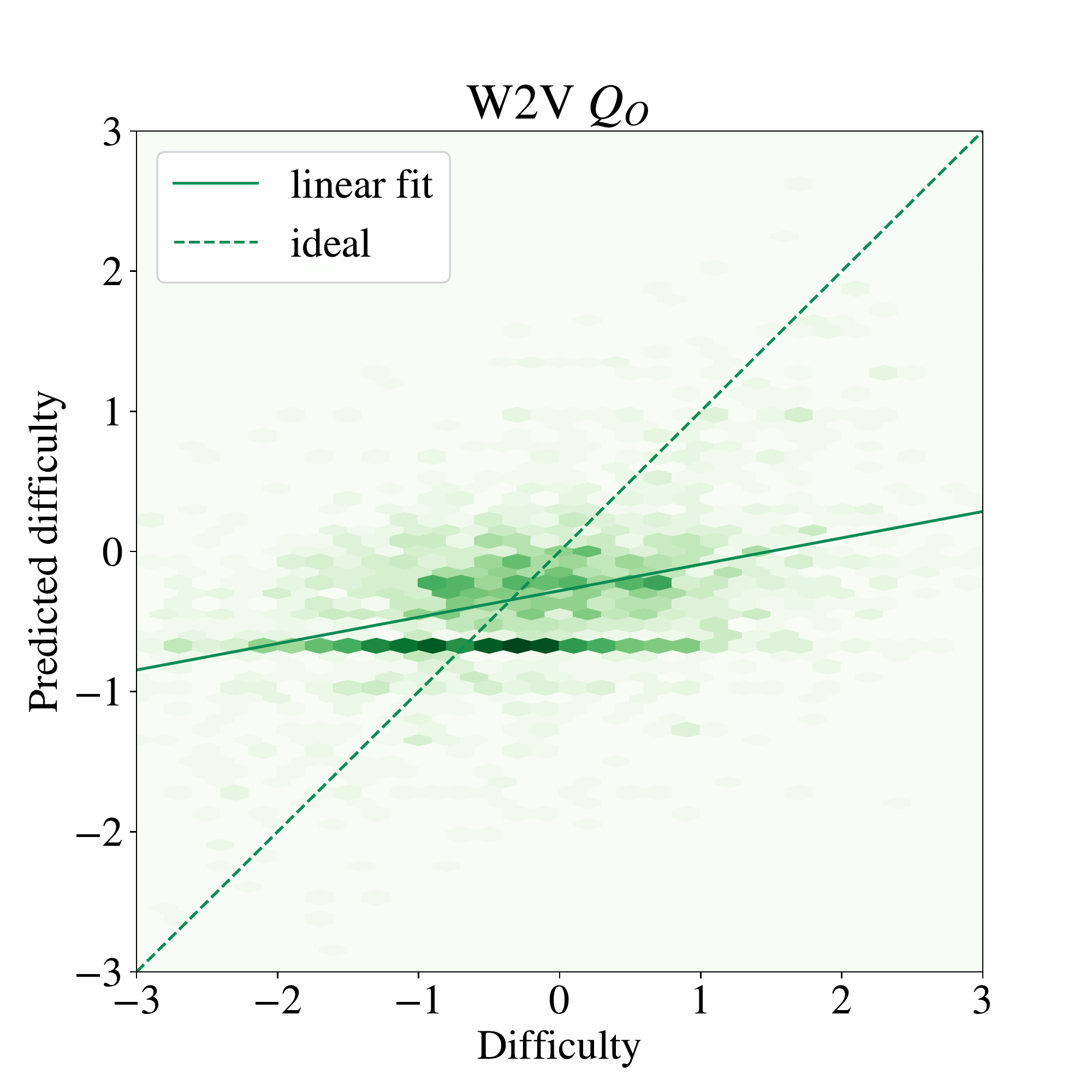}
     \end{subfigure}

       \caption{Distribution of estimated difficulties by true difficulty, \am{}.}
        \label{fig:am_distr_est_diff}
\end{figure}

Lastly, Figure \ref{fig:am_distr_est_diff} plots the distribution of estimated difficulties for \am{}.
Since the difficulties are continuous, we use a density plot, with the true difficulty on the horizontal axis and the estimated difficulty on the vertical axis.
A region with darker color mean that the number of questions in that area is larger.
Again, we plot the linear fit and the ideal (dashed) line.
Although the format is very different to the previous graphs, the findings are similar to \arc{}: although with limitations, BERT and DistilBERT estimate difficulty values across the whole range, while \wv{} often estimates the average difficulty.

\section{Conclusions}
In this paper, we carried out a quantitative study of previous approaches to \qdet{}, to understand how they compare to each other on three publicly available datasets from different educational domains.
We have found that Transformer-based models consistently outperform the other approaches, and that there is not always a large advantage of BERT over DistilBERT, suggesting that in many practical applications it might make sense to use the smaller (and cheaper) distilled model.
Previous research showed that for content knowledge assessment it might also be possible to further improve the \qdet{} performance of Transformers with an additional domain-specific pretraining \cite{benedetto2021application}, but we do not evaluate that possibility as we do not have access to data in the required format.
By training the models on datasets of incremental sizes, we observed that Transformers are better even on smaller datasets (although we never use datasets with less than 4,000 questions).
As for the other approaches, we observed that linguistic features perform well on readability comprehension MCQs, but are not effective on questions from content knowledge assessment, for which TF-IDF and \wv{} embeddings perform better.
%
%
We observed that the imbalance in difficulty labels might affect the estimation, suggesting that it might be worth to focus on finding a trade-off between balancing the difficulty classes and having enough questions for training the models.
Also, future work might focus on deepening the analysis of how the models perform on different difficulty levels, possibly employing different techniques for mapping from continuous estimation to discrete levels and scale linking to improve the estimation accuracy across the whole range of difficulties.

\bibliographystyle{splncs04}
\bibliography{bib}


\end{document}